\documentclass[runningheads]{llncs}
\usepackage{graphicx}
\usepackage{amsmath,amssymb} 
\usepackage{color}
\usepackage[width=122mm,left=12mm,paperwidth=146mm,height=193mm,top=12mm,paperheight=217mm]{geometry}

\usepackage[caption=false]{subfig}
\usepackage{multirow}
\usepackage[american]{babel}

\newcommand{\squishlist}{
	\begin{list}{$\bullet$}
		{ \setlength{\itemsep}{0pt}
			\setlength{\parsep}{3pt}
			\setlength{\topsep}{3pt}
			\setlength{\partopsep}{0pt}
			\setlength{\leftmargin}{1.5em}
			\setlength{\labelwidth}{1em}
			\setlength{\labelsep}{0.5em} } }
	
	\newcommand{\squishend}{
\end{list}  }

\newcommand{\dx}{\,\mathrm{d}x}
\newcommand{\dy}{\,\mathrm{d}y}
\newcommand{\etal}{\textit{et al}.}
\newcommand{\ie}{\textit{i}.\textit{e}.}

\begin{document}
\pagestyle{headings}
\mainmatter
\def\ECCV18SubNumber{***}  

\title{Learning Effective Binary Visual Representations with Deep Networks}



\author{Jianxin Wu and Jian-Hao Luo}
\institute{National Key Laboratory for Novel Software Technology\\Nanjing University, Nanjing, China\\
{\tt\small wujx2001@nju.edu.cn, luojh@lamda.nju.edu.cn}}

\maketitle

\begin{abstract}
Although traditionally binary visual representations are mainly designed to reduce computational and storage costs in the image retrieval research, this paper argues that binary visual representations can be applied to large scale recognition and detection problems in addition to hashing in retrieval. Furthermore, the binary nature may make it generalize better than its real-valued counterparts. Existing binary hashing methods are either two-stage or hinging on loss term regularization or saturated functions, hence converge slowly and only emit soft binary values. This paper proposes Approximately Binary Clamping (ABC), which is non-saturating, end-to-end trainable, with fast convergence and can output true binary visual representations. ABC achieves comparable accuracy in ImageNet classification as its real-valued counterpart, and even generalizes better in object detection. On benchmark image retrieval datasets, ABC also outperforms existing hashing methods.
\end{abstract}

\section{Introduction}

Representation learning, as an alternative name for deep learning, emphasizes the fact that the success of deep learning algorithms hinges on its ability to learn effective representations from data, in particular from a huge number of data instances. In the field of computer vision, the learned deep visual representations are, however, mostly in the format of real-valued vectors or tensors. In this paper, we argue that deep binary visual representations have attractive properties, such as better generalization ability than its floating-point counterparts. Furthermore, when learned appropriately (such as using the method proposed in this paper), binary codes can achieve comparable or even better accuracy than real-valued ones in various vision tasks.

In the vision community, binary representations are mostly preferred for computational reasons. For example, image retrieval is probably the most widely applied domain for binary codes (\emph{i.e.}, binary visual representations). When there are millions or even billions of database images, storing each image as short binary codes (\emph{e.g.}, 48 bits per image) makes possible the storage of all images and on-the-fly searching in the database.

In deep CNN based hashing methods, two-stage methods (\emph{e.g.},~\cite{Xia14AAAI_r11}) were rapidly replaced by single-stage end-to-end learning frameworks. In order to emitting binary codes, often an additional loss term has to be exerted for enforcing binary values~\cite{Liu16CVPR_r12}. Or, some \emph{saturated} functions (such as sigmoid~\cite{Lin15CVPRW_r13} or scaled hyperbolic tangent~\cite{Li17ACMMM_r14,Cao17ICCV_r15}) are applied to the real-valued visual representations. By encouraging these functions to saturate, their output values are \emph{approximately} binary. Saturated functions have been successful in achieving high quality binary codes in CNN-based hashing. However, their drawbacks are also obvious.

First, analogous to the case of ReLU vs. sigmoid as activation functions~\cite{ReLU}, saturated functions are disadvantageous in deep network learning, often leads to slow convergence rate~\cite{Liu16CVPR_r12}. A \emph{non-saturating} activation function or network layer that produces binary values with \emph{fast convergence} is preferred. Second, saturated functions are continuous functions and only produce approximately binary values (such as 0.99 and 0.01 instead of 1 and 0). As for binary visual representations, \emph{true binary values} (\emph{i.e.}, only 0/1 or $\pm 1$ values) are more attractive. Third, hashing methods (including these saturated function based ones) often work at a scale that is much smaller than that of visual recognition problems (such as the ImageNet recognition problem). We want binary visual representations to contribute to \emph{large scale} problems.

Finally and most importantly, image retrieval is only one application of binary codes, being able to generalize and apply to \emph{diverse applications} such as object detection will greatly enrich the applicability of binary codes.

\begin{figure*}[t]
	\centering
	\includegraphics[width=0.98\textwidth]{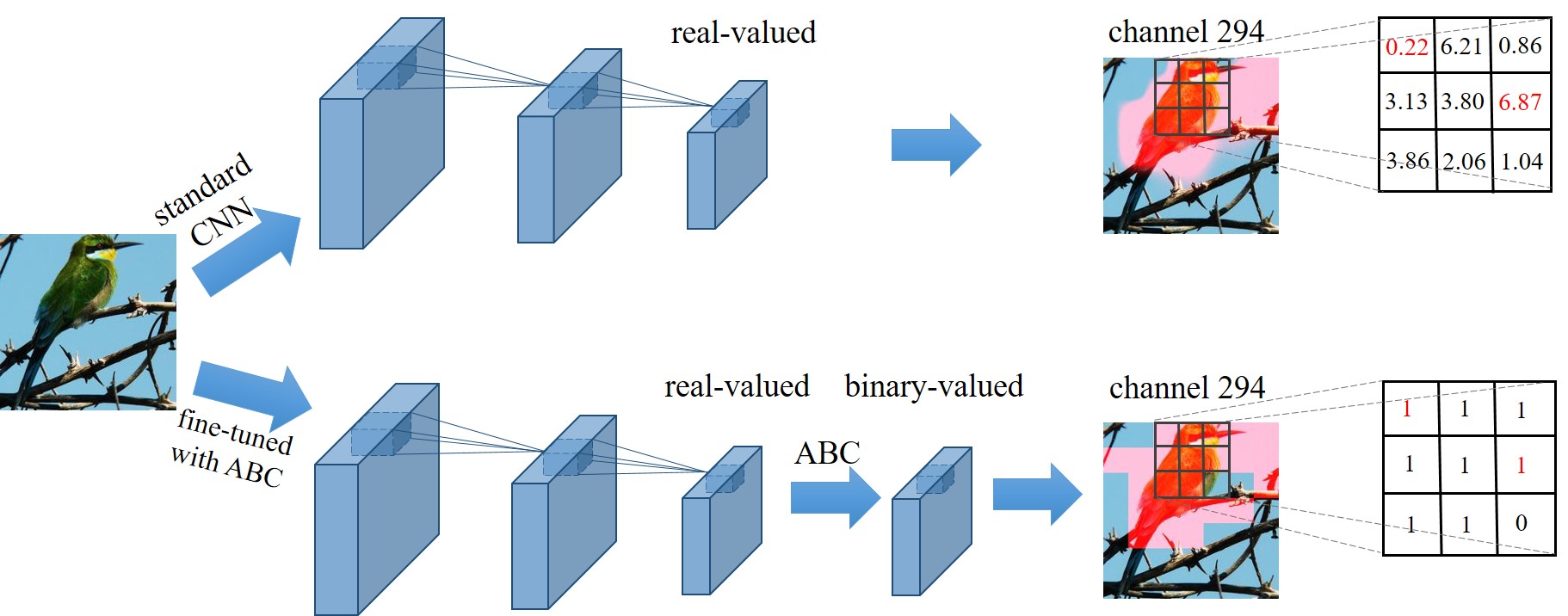}
	\caption{Binary codes (bottom half, produced by the proposed method) \emph{may} achieve better generalization ability than real-valued activation values (top half, produced by ResNet-50), whose magnitudes may be misleading.} \label{fig:motivation}
\end{figure*}

Fig.~\ref{fig:motivation} inspires our thirst for binary visual representation beyond the retrieval task. When the input image is fed into two different networks, the top one is ResNet-50~\cite{ResNet} with real-valued outputs and the bottom one is proposed in this paper, which outputs binary codes with an approximately binary clamping (ABC) layer.

For the last convolution block (the one before global average pooling), we randomly picked a convolution channel (294th) and the pink regions are those that are activated (channel output $>0$). Now observing the region covered in the $3 \times 3$ grid, the two activation values in red for ResNet-50 (with values 0.22 and 6.87) vary greatly in terms of their magnitude. However, they are both borderline regions that contain small part of the object but mostly background. In other words, activation magnitudes are not necessarily important, what is important is whether it is activated or not. In short, Fig.~\ref{fig:motivation} indicates that \emph{a binary (\ie, activated or not) visual representation may have better generalization} than real-valued features, as shown in the bottom half of this figure.

Thus, in this paper, we propose Approximately Binary Clamping (ABC), a novel activation function that can produce both approximately and true binary output values depending on its parameter value. The key advantages and major contributions of this paper are summarized as follows.

\begin{itemize}
	\item ABC is non-saturating and has very fast convergence speed in practice, which can emit true binary values. For short binary codes, it achieves better or comparable results than existing hashing methods in image retrieval tasks.
	
	\item However, image retrieval (hashing) is only one application of binary codes. In this paper, we would show that approximately or true binary representations can achieve comparable or even better accuracy than its real-valued counterparts in various vision tasks, such as large-scale image recognition or object detection. In other words, exploring the generalization ability of binary representation is our novel contribution, which has not been proposed or explored in previous works.
	
	\item On ImageNet classification, when ABC is added to ResNet-50 (a state-of-the-art method) to produce true binary visual representations (2048 0/1 values vs. 2048 floating point ones in ResNet-50), its accuracy is only 0.298\% below ResNet. That is, it is suitable for large scale problems. To the best of our knowledge, this is the first binary representation for a large-scale problem with comparable accuracy.
	
	\item The learned codes are generalizable. We applied the learned model to object detection, and the ABC model obtains roughly 0.5--1\% higher detection mAP than ResNet-50.
	
	\item ABC is a piecewise linear activation layer and end-to-end trainable, which takes little computation. It does not require additional loss term to enforce true binary values.
\end{itemize}

\section{Related Work}

Binary representations are closely related to hashing method, which aim at producing a compact binary code for approximately nearest neighbor search. Hashing for image retrieval has been widely studied in the community. We can roughly divide these hashing methods into two categories: unsupervised and supervised.

Typical unsupervised hashing uses reconstruction error minimization~\cite{Gionis99VLDB_LSH,Gong11CVPR_r1,Jegou11PAMI_r2} or graph learning~\cite{Weiss14NIPS_r3,Liu11ICML_r4,Liu14NIPS_r5} to encode unlabeled points into binary codes without any supervision. One representative approaches is Locality Sensitive Hashing (LSH)~\cite{Gionis99VLDB_LSH}. LSH projects the data points into random hyperplanes. However, LSH requires long code length to obtain satisfactory retrieval performance. Another representative method is Spectral Hashing (SH)~\cite{Weiss14NIPS_r3}, which minimizes the weighted Hamming distance of image pairs. 

Although some of these unsupervised hashing methods can be adapted to suit the supervised settings, they often cannot effectively make good use of data points and their labels. Hence, more attentions have been focused on supervised hashing~\cite{Liu12CVPR_r6,Norouzi11ICML_r7,Shen15CVPR_r8} to deal with more complicated issues.

To generate effective hash codes, supervised methods usually use pairwise constraints, \ie, two data points are similar if they belong to the same class, and vice versa. After training, similar data points should be mapped to similar binary codes in the Hamming space. However, it may require a large sparse matrix to describe the image similarity, which raises serious issues when dealing with large-scale datasets.

In the past few years, the development of deep neural networks have advanced the state-of-the-art in many computer vision tasks~\cite{AlexNet,ResNet}, including image retrieval. Researchers have explored and proposed numerous deep learning based hashing methods and algorithms to improve the retrieval accuracy~\cite{Salakhutdinov09_r9,Torralba08CVPR_r10,Xia14AAAI_r11,Liu16CVPR_r12,Lin15CVPRW_r13,Li17ACMMM_r14,Cao17ICCV_r15}. The earliest work in these methods may be Semantic Hashing~\cite{Salakhutdinov09_r9}, which  builds a graphical model by using a Restricted Boltzmann Machine (RBM). However, Semantic Hashing is designed for document retrieval. To explore the performance of deep RBM on image datasets, a supervised deep RBM is proposed in~\cite{Torralba08CVPR_r10}.

CNN-based hashing methods have recently been widely studied, showing impressive results on image retrieval. Xia~\etal~\cite{Xia14AAAI_r11} proposed a two stage hashing approach to learn feature representation as well as binary hash codes. In contrast, Liu~\etal~\cite{Liu16CVPR_r12} combined these two stages together in an end-to-end framework. Lin~\etal~\cite{Lin15CVPRW_r13} explored a sigmoid activated encoding layer to generate the approximately binary hash codes using a standard classification loss. This framework is simple but efficient, and can learn a set of effective hash-like functions on different challenging datasets. However, sigmoid function can not generate a real binary code, which may hurt the final retrieval results. Hence, Li~\etal~\cite{Li17ACMMM_r14} and Cao~\etal~\cite{Cao17ICCV_r15} introduced a scaled hyperbolic tangent activation function $h(\mathbf{x})=\tanh(\alpha \mathbf{x})$ with $\alpha \ge 1$ to closely approximate the sign function when $\alpha$ is large. This idea is similar to our ABC method. However, $\tanh(\alpha x)$ is a saturated function which only outputs \emph{soft} binary values even when $\alpha$ is large. In contrast, ABC is non-saturating and can produce true binary values.

Another research line to produce binary representations using deep learning methods are those binarized networks, which are designed for fast inference and small memory footprint~\cite{Courbariaux15NIPS,Hubara16NIPS_Binary,Rastegari16ECCV,Li17ICCV}, in which binary visual representation is only a by-product. It is worth noting that the accuracies of binarized neural networks are often lagging behind their real-valued counterparts by a very large margin, because these networks binarize all layers. The visual representation by the proposed ABC layer, as we will show, only binaries the visual representation itself and achieves comparable or even higher accuracies in various tasks than real-valued networks.

\section{Approximately Binary Clamping}

In this section, we propose the approximately binary clamping (ABC) activation function, and compare it with scaled hyperbolic tangent.

\subsection{The ABC layer}

\begin{figure*}[t]
	\centering
	\subfloat[ABC] { \includegraphics[width=0.45\textwidth]{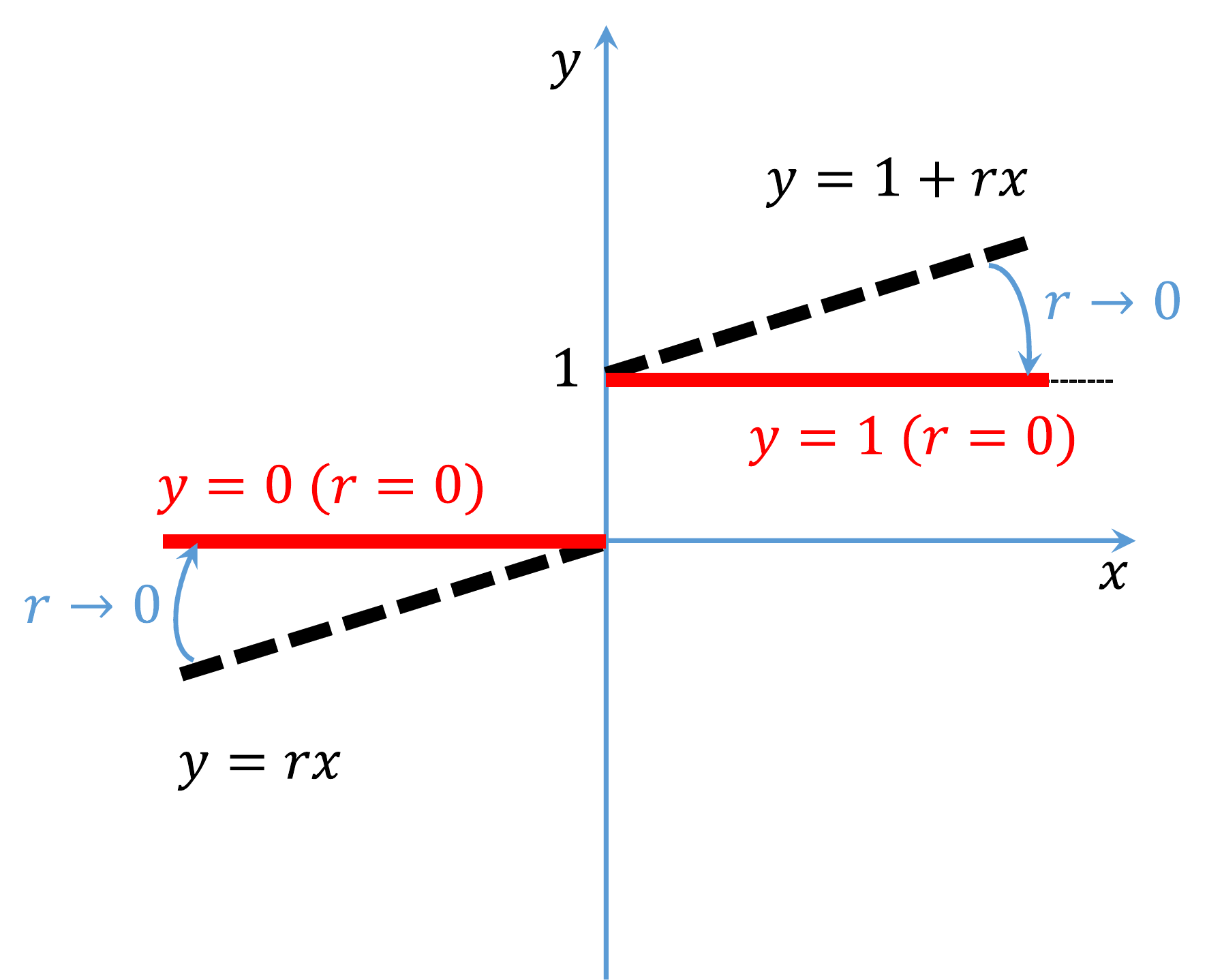} \label{fig:ABC} }
	\hspace{12pt}
	\subfloat[$\tanh(\alpha x)$: Scaled $\tanh$] { \includegraphics[width=0.45\textwidth]{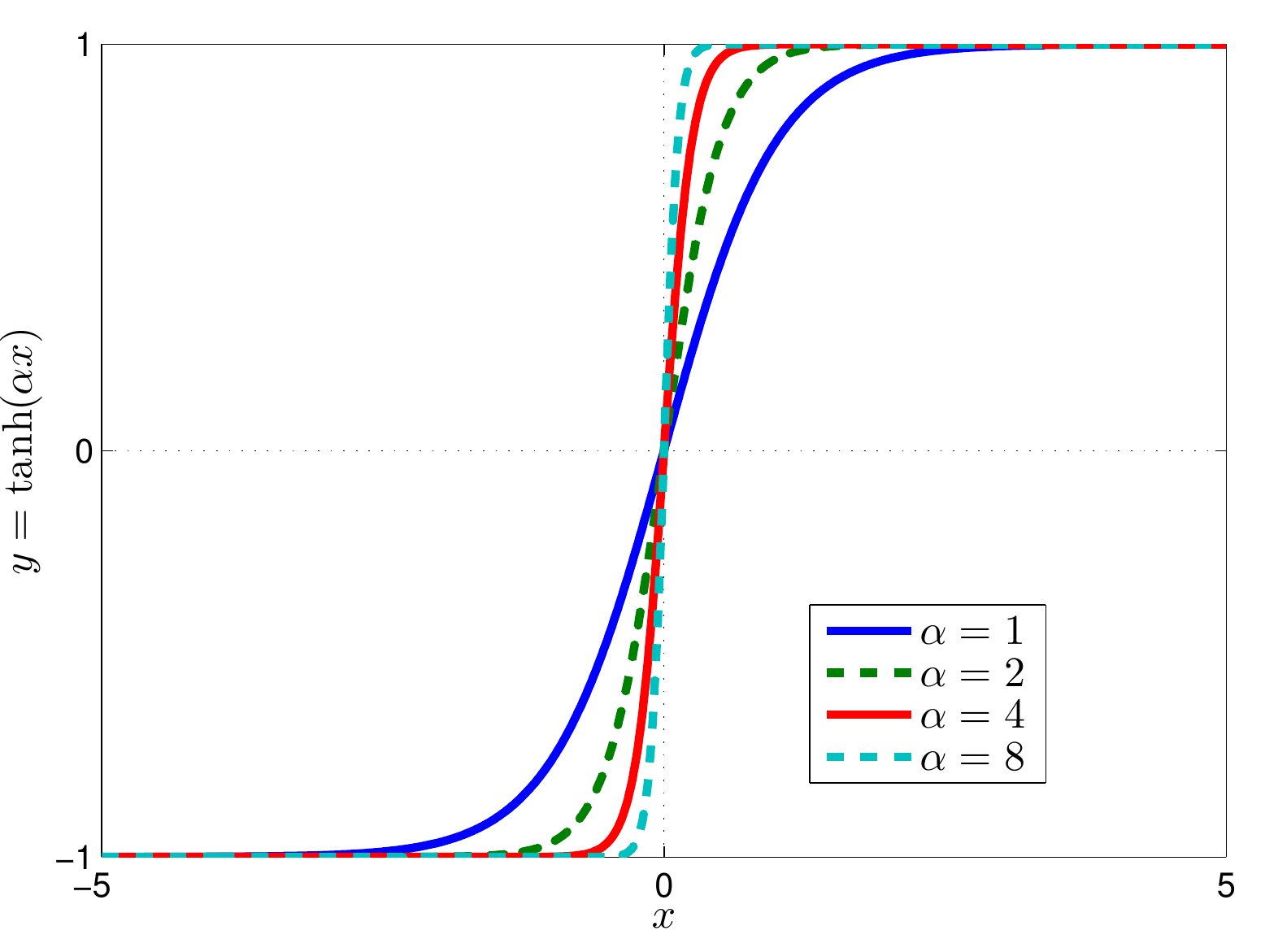} \label{fig:Tanh} }
	\caption{Illustration of the proposed ABC activation function and the scaled $\tanh$ function.} \label{fig:ABC_Tanh}
\end{figure*}

The proposed ABC activation function applies element-wise to all elements in a tensor. For $x \in \mathbb{R}$, its output $y$ is defined as 
\begin{equation}
y = \left\{ 
\begin{aligned}
1 + rx &&& \text{if $x > 0$} \\
rx &&& \text{otherwise}
\end{aligned}
\right. \,,
\end{equation}
in which $r \ge 0$ is a parameter. As illustrated in Fig.~\ref{fig:ABC}, ABC is a simple piecewise linear function. Similar to ReLU, ABC is not continuous at $x=0$. However, for all other values, we have 
\begin{equation}
\frac{\dy}{\dx} = r \,,
\end{equation}
and we assume $\frac{\dy}{\dx}=r$ when $x=0$. This simple gradient rule not only makes ABC end-to-end trainable in deep neural networks, but also guarantees that it has negligible computational cost in both forward and backward computations.

A few special values for $r$ are worth observing.
\begin{itemize}
	\item When $r=1$, ABC is like the identity function $y=x$, except that positive values are elevated by 1.
	\item When $0<r<1$, as shown by the black dashed line segments in Fig.~\ref{fig:ABC}, ABC \emph{shrinks} the gap between $y$ and the target binary values (0/1). However, ABC is non-saturating. For $x$ with large magnitudes ($|x| \gg 1$), $y$ can still be far away from binary.
	\item On the other hand, since we add a batch normalization (BN) layer~\cite{bn} before the ABC layer (\emph{i.e.}, $x$ is the output of BN), we would expect the average scale of $x$ to remain stable or gradually reduce when we gradually reduce the value of $r$ towards 0 in different training epochs. Then, $|rx| \rightarrow 0$ when $r \rightarrow 0$, as illustrated by the two blue arcs in Fig.~\ref{fig:ABC}. In other words, $y$ is approaching binary values when $r$ approaches 0, explaining ``approximately binary'' in the name.
	\item When $r=0$, $y$ becomes \emph{true} binary values, as illustrated by the two red line segments in Fig.~\ref{fig:ABC}. In other words, when $r$ gradually reduces and finally reaches 0, the output of ABC is ``clamped'' to be binary.
\end{itemize}

Note that the gradient of the ABC layer would be $0$ when $r=0$. So all the layers before ABC will not be updated. However, the training procedure will continue. For example, in the image recognition task, the layers after ABC (whose input is the learned binary representations) are still being trained via back-propagation such that better classifiers can be learned. Since we only set $r=0$ in very late epochs (e.g., starting from the 17th out of 20 epochs), visual representations are already stabilized at that time and setting $r=0$ will not hurt the network's accuracy. Also note that in $\tanh(\alpha x)$ based methods, $\alpha$ is large in the final epochs and its gradient is almost 0 everywhere, too.

We implemented ABC in both Torch and Caffe.\footnote{Code will be released.} As will be shown in the experiments, by adding a BN layer and an ABC layer after one target layer, we can force the target layer to emit binary codes by a suitable schedule of $r$ value sequences in the ABC layer. One interesting observation is that adding a BN before ABC is crucial for its success. Without BN, the transition to a smaller $r$ can be unstable. Hence, we use the BN+ABC combination in all our experiments.

\subsection{ABC vs. scaled $\tanh$}

Scaled hyperbolic tangent is a recently popularized activation function to obtain binary codes, which is defined as
\begin{equation}
y = \tanh(\alpha x) \,,
\end{equation}
in which $\alpha>0$ is a scaling parameter, whose role is similar to the $r$ parameter in our ABC layer.

As shown in Fig.~\ref{fig:Tanh}, as $\alpha$ gets bigger, the $\tanh(\alpha x)$ curve approaches the two binary values ($\pm 1$ instead of 0/1, but this difference is trivial). When $\alpha$ is around 10, $\tanh(\alpha x)$ already emits approximately binary values, as illustrated by the light blue curve with $\alpha=8$. Thus, by gradually increasing $\alpha$ from 1 to a large value (such as 10), $\tanh(\alpha x)$ can emit approximately binary values, too.

However, $\tanh(\alpha x)$ will \emph{never} be true binary. Even for a very large $\alpha$ value, \emph{e.g.}, $\alpha=10000$ (which is not practical in real networks), a very small $x$ could still generate $y$ that is far from being binary. For example, when $\alpha=10000$ and $x=0.0001$, $\tanh(\alpha x)=0.7616$, far away from both $-1$ and $+1$. Although the chance of a very small $x$ is not high, very large $\alpha$ value is not suitable for training deep neural networks. In our experiments, when using $\alpha=10000$, our CNN learning process almost immediately fell into numerical errors and activation values became NaN.

Another drawback for $\tanh(\alpha x)$, beyond being only \emph{soft binary}, is that it slows down convergence~\cite{Liu16CVPR_r12}, which makes it not suitable for large scale problems because many gradient values will be 0~\cite{ReLU} even when $\alpha=1$.

On the contrary, ABC has nicer gradient properties than $\tanh(\alpha x)$. Because the gradient is always $r$, when $r=1$ or $r$ is relatively large, the ABC layer will not hinder gradient back-propagation or network training. When we gradually reduce $r$ by $k$ times, layers after ABC will not be affected by this change, but any layer before ABC will observe its gradient's magnitude reduced by $k$ times. Since in training deep networks we reduce the learning rate regularly, the effect of ABC to previous layers can be easily mitigated by the following trick for large scale problems: if in regular training without the ABC layer, we reduce the learning rate by $k$ times, then instead we reduce both $r$ and the learning rate by $\sqrt{k}$ times when ABC is used.

This trick works particularly well for fine-tuning a pretrained model. A commonly used trick in fine-tuning is to use smaller learning rates for early layers, but a large learning rate for newly initialized ones (\emph{e.g.}, the last fully connected layer). When fine-tuning a pretrained network with ABC, we can set $r<1$, which automatically guarantees that layers before ABC will implicitly have smaller learning rates than those after ABC.

\section{Experimental Results}

In this section, we empirically illustrate the benefits of the proposed ABC activation function and layer in three types of task: learning short binary codes for image retrieval (Sec.~\ref{sec:exp:hashing}), learning long binary visual representations for large scale image recognition (Sec.~\ref{sec:exp:imagenet}) with a convergence study (Sec.~\ref{sec:exp:convergence}), and generalizing the learned recognition network to object detection (Sec.~\ref{sec:exp:detection}).

\subsection{Hash learning for image retrieval} \label{sec:exp:hashing}

In this part, we compare the performance of our ABC method with a recently proposed activation function: scaled $\tanh$, which is introduced by Li~\etal~\cite{Li17ACMMM_r14} and Cao~\etal~\cite{Cao17ICCV_r15}. Both of them employed a scaling parameter $\alpha$ in the hyperbolic tangent function: $h(\mathbf{x})=\tanh(\alpha \mathbf{x})$ to control the smoothness of the original function. By changing the value of $\alpha$, the scaled $\tanh$ function would become more smooth or more saturated. When $\alpha$ is large enough, this function would approach the standard sign function. The major difference between~\cite{Li17ACMMM_r14} and~\cite{Cao17ICCV_r15} is that: Li~\etal~\cite{Li17ACMMM_r14} let the network itself to learn an appropriate scale value, while Cao~\etal~\cite{Cao17ICCV_r15} gradually increased this value to force the network to learn a more hash-like code. In our experiments, \cite{Cao17ICCV_r15} is chosen as the baseline to compare with our ABC method, because it is conceptually simpler and reported higher mAP than the method in~\cite{Li17ACMMM_r14}. We will refer to the method in~\cite{Cao17ICCV_r15} as adaptive $\tanh$.

For fairness of comparison, we use the same CNN structure introduced in DSH~\cite{Liu16CVPR_r12}. This model consists of 3 convolution-pooling layers and 2 fully connected layers to learn the similarity of different image pairs. We append the activation layer (ABC or scaled $\tanh$) in the last fully connected layer, and replace the loss function if necessary. We use the source code of DSH\footnote{\url{https://github.com/lhmRyan/deep-supervised-hashing-DSH}} to train the networks, which means binary codes are all learned within the same framework. Two widely used benchmarks are utilized for evaluation, including CIFAR-10~\cite{cifar_10} and NUS-WIDE~\cite{NUS_WIDE}. We summarize these two datasets as follows:

\begin{itemize}
	\item \textbf{CIFAR-10: }This dataset consists of 10 categories and totally 60,000 $32\times 32$ object images. We use the official train/test split in our experiments, \ie, we learn the binary codes on 50,000 training images and evaluate the performance on 10,000 test images. 
	\item \textbf{NUS-WIDE: }We follow the DSH setting~\cite{Liu16CVPR_r12} to use a subset of NUS-WIDE dataset here. There are totally 21 most frequent concepts, and 195,834/10,000 $64\times 64$ resized images are used for training/evaluation, respectively. NUS-WIDE is a multi-label dataset, and two images are considered as similar to each other if they share at least one common positive label.
\end{itemize}

\begin{table*}[!t]
	\caption{\textbf{Comparisons of different hashing methods (evaluated by mAP values).} We divided these published hashing algorithms into three groups: traditional methods, CNN-based methods and the modified DSH approach with different activation functions. Note that, Liu \etal have updated their DSH method, hence the results are higher than those reported in their original publication. All the CNN-based methods are trained using the structure introduced in~\cite{Liu16CVPR_r12}.}
	\label{Table1} 
	\begin{center}
		\begin{tabular}{l||c|c|c|c||c|c|c|c}
			\hline
			\multirow{2}{*}{Method} & \multicolumn{4}{c||}{CIFAR-10} & \multicolumn{4}{c}{NUS-WIDE} \\
			\cline{2-9} & 12-bits & 24-bits & 36-bits & 48-bits & 12-bits & 24-bits & 36-bits & 48-bits \\
			\hline
			LSH~\cite{Gionis99VLDB_LSH} & 0.1277 & 0.1367 & 0.1407 & 0.1492 & 0.3329 & 0.3392 & 0.3450 & 0.3474 \\
			SH~\cite{Weiss14NIPS_r3} & 0.1319 & 0.1278 & 0.1364 & 0.1320 & 0.3401 & 0.3374 & 0.3343 & 0.3332 \\
			CCA-ITQ~\cite{Gong11CVPR_r1} & 0.1653 & 0.1960 & 0.2085 & 0.2176 & 0.3874 & 0.3977 & 0.4146 & 0.4188 \\
			KSH~\cite{Liu12CVPR_r6} & 0.2948 & 0.3723 & 0.4019 & 0.4167 & 0.4331 & 0.4592 & 0.4659 & 0.4692 \\
			\hline
			CNNH~\cite{Xia14AAAI_r11} & 0.5425 & 0.5604 & 0.5640 & 0.5574 & 0.4315 & 0.4358 & 0.4451 & 0.4332 \\
			DLBHC~\cite{Lin15CVPRW_r13} & 0.5503 & 0.5803 & 0.5778 & 0.5883 & 0.4663 & 0.4728 & 0.4921 & 0.4916 \\
			DNNH~\cite{Lai15CVPR_r16} & 0.5708 & 0.5875 & 0.5899 & 0.5904 & 0.5471 & 0.5367 & 0.5258 & 0.5248 \\
			DSH~\cite{Liu16CVPR_r12} & 0.6157 & 0.6512 & 0.6607 & 0.6755 & 0.5483 & 0.5513 & 0.5582 & 0.5621 \\
			SUBIC~\cite{SUBIC} & 0.6349 & 0.6719 & 0.6823 & 0.6863 & -- & -- & -- & -- \\
			\hline
			DSH-new w/ BN & 0.6780 & 0.6927 & 0.7040 & 0.7115 & 0.5795 & 0.5949 & 0.5963 & 0.6018 \\ 
			DSH-new w/o BN & 0.6326 & 0.6812 & 0.6981 & 0.7177 & 0.5849 & \textbf{0.6020} & \textbf{0.6075} & \textbf{0.6073} \\
			$\tanh(\alpha x)$ w/ BN & 0.6799 & 0.6908 & 0.6974 & 0.6974 & 0.5796 & 0.5987 & 0.6026 & 0.5959 \\
			$\tanh(\alpha x)$ w/o BN & 0.6397 & 0.6887 & 0.7042 & 0.7097 & \textbf{0.5906} & 0.5942 & 0.5977 & 0.6007 \\
			ABC with BN (ours) & \textbf{0.6881} & \textbf{0.7149} & \textbf{0.7193} & \textbf{0.7201} & 0.5882 & 0.5920 & 0.5993 & 0.5978 \\
			\hline
		\end{tabular}
	\end{center}
\end{table*}

Following the experimental setting of the new DSH method, we first train a standard DSH network with 12 bits as the base model. Starting from this model, we then replace the 12 bits layer with a larger one (24 bits, 36 bits and 48 bits). The new added fully-connected layer is initialized using the Xavier initialization method. Because the adaptive $\tanh$ released code is in the Caffe framework,\footnote{\url{https://github.com/thuml/HashNet}} we use Caffe for ABC too.

We use SGD (stochastic gradient descent) with 0.9 momentum and 0.004 weight decay to fine-tune the new model in 30,000 iteration (batch size is 200). The initial learning rate is set to $10^{-4}$ and multiplied by 0.6 after every 4000 iterations. To be consistent with DSH, we set $\alpha=0.01$, $m=2k$ in the loss function. The scaling parameter $\alpha$ of adaptive $\tanh$ is updated according to $$\alpha=(1+0.005i)^{0.5} \,,$$ in which $i$ is the iteration index. As for ABC, the parameter $r$ is initialized by 1, multiplied by 0.95 after each epoch, and finally the minimum value is 0.002. On the NUS-WIDE dataset, $r$ is multiplied by 0.94 (1000 iterations per epoch), and stops at the minimum value $r=0.1662$. And, we find that if we fix the learning rate as $10^{-4}$, we could obtain higher mAP results for all these methods on NUS-WIDE. Thus, we do not reduce the learning rate when experimenting with NUS-WIDE.

After training, the output of ABC and adaptive $\tanh$ are both approximately binary representations. In order to generate true binary codes, parameter $r$ is set to $0$ when we use the trained ABC model to extract hash code. Then, all the $0$s are replaced with $-1$s, hence the $\pm 1$ values can be used for image retrieval. As for adaptive $\tanh$ method, we use signum function to get a final representation. And we assume $\text{sgn}(x)=1$ when $x=0$, in order to get a true binary code. These binary features are finally evaluated (mAP values) using the codes provided by official DSH project.

We show the comparison results in Table~\ref{Table1}. The first two groups of rows represent traditional hashing methods and CNN-based methods respectively. As these results clearly indicate, CNN-based methods consistently outperform traditional non-deep-learning algorithms, and in most cases by a large margin. Thanks to the strong generalization ability of representation learning, the CNN-based methods achieve much better performance. We also want to note that DSH is not a very powerful CNN architecture. If a more powerful CNN model is adopted, such as the residual network~\cite{ResNet}, retrieval results could be further improved for CNN-based methods.

We then compare our ABC method with another similar function: adaptive $\tanh$. We append the activation function after the last fully connected layer, and use the DSH framework to train the model. Since our ABC needs the batch normalization (BN)~\cite{bn} layer (only included in the last layer, \ie, before the ABC activation function) to prevent activation values from increasing when $r$ decreases, we also report the performance of DSH with/without BN (the original DSH paper did not use BN). Note that, Liu \etal~have updated their DSH method, hence the results generated by the new code are higher than the original reported values. To distinguish between these two results, we use DSH-new to denote the new framework (also used in our comparisons), and let DSH to represent the original publication (in the second group in Table~\ref{Table1}).

An interesting result is that, there is no obvious winner between using BN (``w/ BN'' in Table~\ref{Table1}) or not (``w/o BN'' in Table~\ref{Table1}). On the CIFAR-10 dataset, using BN could improve retrieval performance with short codes (12 bits or 24 bits). However, when the code length is longer, it may hurt retrieval mAP. Similar phenomena could also be observed in the adaptive $\tanh$ method. Although adaptive $\tanh$ explicitly requires binary values while DSH only encourages binary values through an implicit loss term, the adaptive $\tanh$ method can sometime lead to worse performance than the standard DSH. As indicated in~\cite{Liu16CVPR_r12}, utilizing saturated functions such as \emph{sigmoid} or $\tanh$ would inevitably slow down or even restrain the convergence of the network. 

In contrast, the proposed ABC method shows consistently better retrieval performance than DSH and adaptive $\tanh$ on the CIFAR-10 dataset, no matter whether batch normalization is used or not in these methods. There is no obvious winner between DSH and adaptive $\tanh$, but both are consistently worse than ABC. In a few cases, ABC has a large margin over other methods. For example, with 36 bits, ABC has more than 1.5 percentage point advantage over all 4 competitors.

NUS-WIDE is a multi-label retrieval dataset, which is a complicated task. The five methods (ABC, DSH and adaptive $\tanh$ with or without batch normalization) all have quite similar results. In all bit length settings, ABC lags behind the best method in no more than 1 percentage point. The proposed ABC method yielded comparable but more robust performance even with a short binary code.

One interesting observation on this dataset is that the DSH (new code) method often achieves the highest mAP among all compared methods, better than both ABC and adaptive $\tanh$. One possible cause for this phenomenon may be attributed to the multi-label nature of this dataset. It is interesting to see how explicit binarization methods such as ABC and adaptive $\tanh$ can be improved under the multi-label setting.

\subsection{Large scale recognition} \label{sec:exp:imagenet}

We also compare ABC and adaptive $\tanh$ on the ImageNet recognition task. For ABC, we use the Facebook Torch ResNet-50 pretrained network and code.\footnote{\url{https://github.com/facebook/fb.resnet.torch}} We added a linear layer ($2048 \mapsto 2048$) after the global average pooling layer, then added a BN layer and an ABC layer afterwards, thus producing 2048 binary bits as our visual representation.

Following the Facebook code, we use SGD for optimization. The mini-batch size is 256, momentum (0.9) and weight decay (0.0001) remain unchanged. We fine-tune the pretrained ResNet-50 model for 20 epochs. The fine-tuning initial learning rate is 0.01, and the learning rate is multiplied by $\sqrt{0.1}$ after every 4 epochs. For the ABC layer, the initial $r$ value is 0.1 because this is fine-tuning a pretrained model in a large scale problem. Then, after every 4 epochs $r$ is multiplied by $\sqrt{0.1}$. However, in the last 4 epochs (17--20), $r$ is set to 0 because we want true binary visual representations. In this case, only the last fully connected layer is still being trained, because the gradient of ABC layer would vanish.

For the adaptive $\tanh$ method, we used a similar network architecture. In the pretrained ResNet-50, we added a linear ($2048 \mapsto 2048$), a BN and an adaptive $\tanh$ layer after the global average pooling layer. Since the released adaptive $\tanh$ code is based on Caffe, we use Caffe to utilize the official code for this $\tanh$ experiment.

The pretrained Caffe ResNet-50 network was fine-tuned for 20 epochs using SGD, with batch size 256, momentum 0.9, weight decay 0.005 and initial learning rate 0.001, which are based on recommended practices in Caffe. The learning rate was multiplied by 0.1 after the 10th and 16th epoch, respectively. The $\alpha$ value in the adaptive $\tanh$ function was set as $$\alpha = (1+15e)^{0.4} \,,$$ in which $e$ is the epoch index. Note that $\alpha$ is updated once every 2 epochs. The maximum $\alpha$ value is $\alpha=9.401$, under which $\tanh(\alpha x)$ is a quite good approximation to the sign function. This set of rules for updating $\alpha$ is the best practice we found in our experiments.

Note that if we treat the ABC or $\tanh$ outputs as visual representations, they are 2048 bits and they are the input to the linear classifier layer to classify one image into one of the 1000 categories of ImageNet objects.

Now we analyze the results by comparing different models' accuracy on the validation set. For a fair comparison, the results are compared with their official model, \ie, Torch ResNet-50 for ABC, Caffe ResNet-50 for adaptive $\tanh$. For ABC, its baseline method (Facebook Torch ResNet-50, in which the visual representations are 2048 real-valued numbers) has a top-1 error of 24.018\%. The ABC version (in which the visual representations are only 2048 bits) achieved the top-1 error of 24.316\%. Note that the difference is only 0.298\%!

In other words, when we convert the visual representation from a real-valued vector to a bit string with the same length, we only lost less than 0.3 percentage points in the ImageNet large scale recognition problem. To the best of our knowledge, our proposed ABC layer is the first method that achieves a comparable accuracy as its real-valued counterpart. Although there are binarized networks such as the XNOR net~\cite{Rastegari16ECCV}, their accuracies are much lower than real-valued networks. One big difference between ABC and binarized networks is that ABC only binarize the final visual representation, rather than all layers. It is an interesting direction to explore, though, on how to binarize many even all layers while still preserving the network accuracy.

For adaptive $\tanh$, its baseline method (Caffe ResNet-50 model) has a top-1 error of 24.365\%, which is different from the Torch baseline for ABC, but the two baseline accuracy rates (24.018 vs. 24.365) are quite close to each other. The adaptive $\tanh$ version, however, only achieved a top-1 error of 26.430\%. The difference between the real-valued and binary representation of adaptive $\tanh$ is 2.065 percentage points, which is a significant amount. Note that the visual representation in adaptive $\tanh$ is only soft binary, we expect that its error will slightly increase if we take measures to force them to be true binary.

Comparing the differences between real-valued and binary representations, it is only 0.298\% for ABC and 2.065\% for adaptive $\tanh$. ABC has a clear advantage in this large scale recognition problem over adaptive $\tanh$, which corroborates observations that frequently appear in our community: saturated functions are not friendly to large scale deep learning. Hence, ABC is a good choice for learning binary representations for large scale vision problems.

One note for this section: when we replace the linear layer before ABC to produce 4096 outputs (\emph{i.e.}, the inserted linear layer is now $2048 \mapsto 4096$), ABC can achieve a top-1 error of 24.064\%, almost matches the real-valued baseline (24.016\%). 

Finally, we want to emphasize that, our motivation of this section is not a better image recognition system, but an image representation that generalizes well. CNN features (activation values) vary a lot in scale (e.g., 500 vs. 0.5), and we argue that a binary or approximately binary representation makes us focus on whether a neuron is activated (but not its activation scale). As we have shown above, binary visual representation can also generalize well in large scale problem, which has not been explored in previous works. It is interesting to see whether we can use ABC to improve the recognition accuracy in other dataset. We would explore it in the future.

\subsection{Convergence study} \label{sec:exp:convergence}

\begin{figure*}[t]
	\centering
	\subfloat[ImageNet recognition (the lower the better)] { \includegraphics[width=0.34\textwidth]{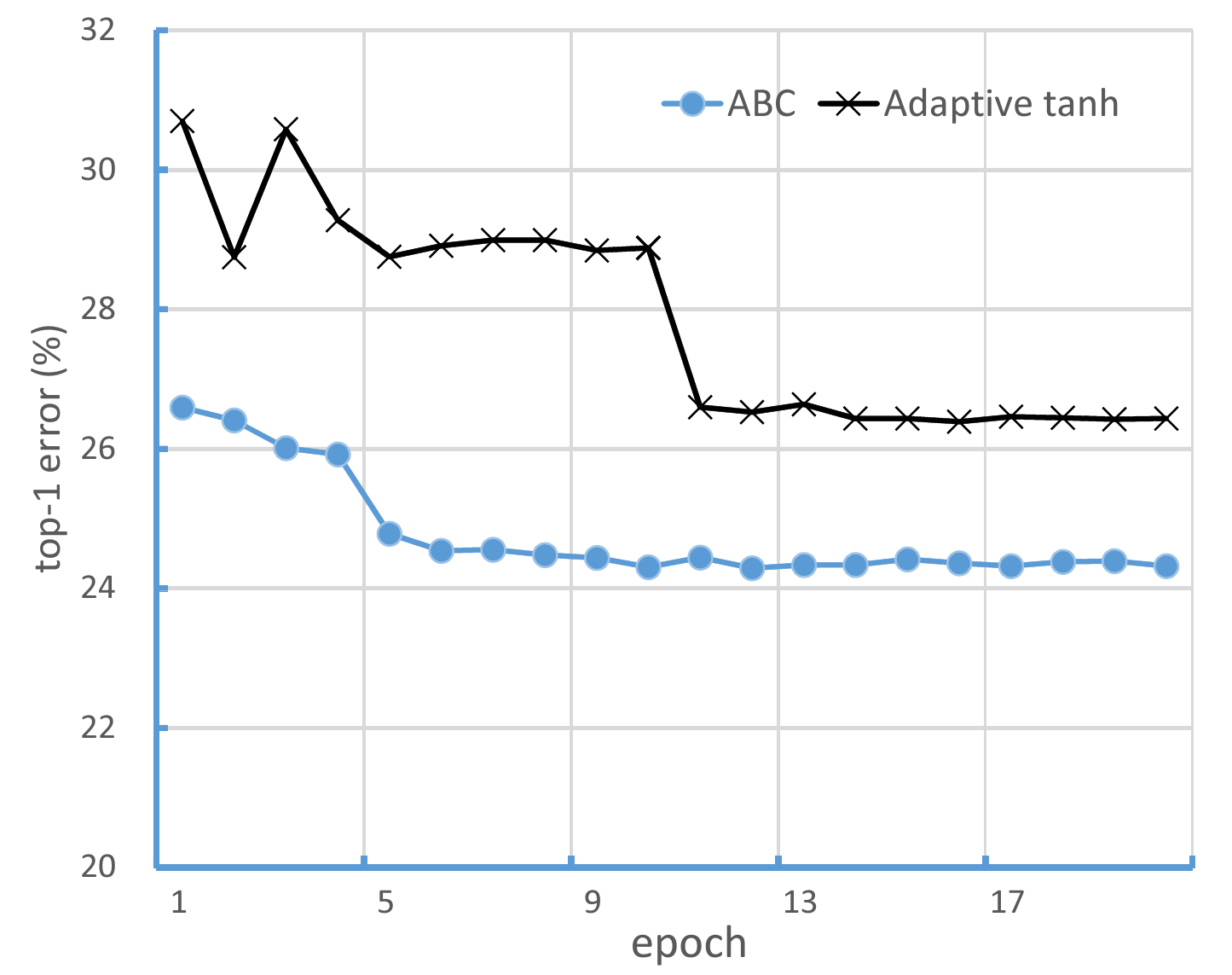} \label{fig:convergence_ImageNet}}
	\hspace{10pt}
	\subfloat[CIFAR-10 retrieval 48 bits (the higher the better)] { \includegraphics[width=0.34\textwidth]{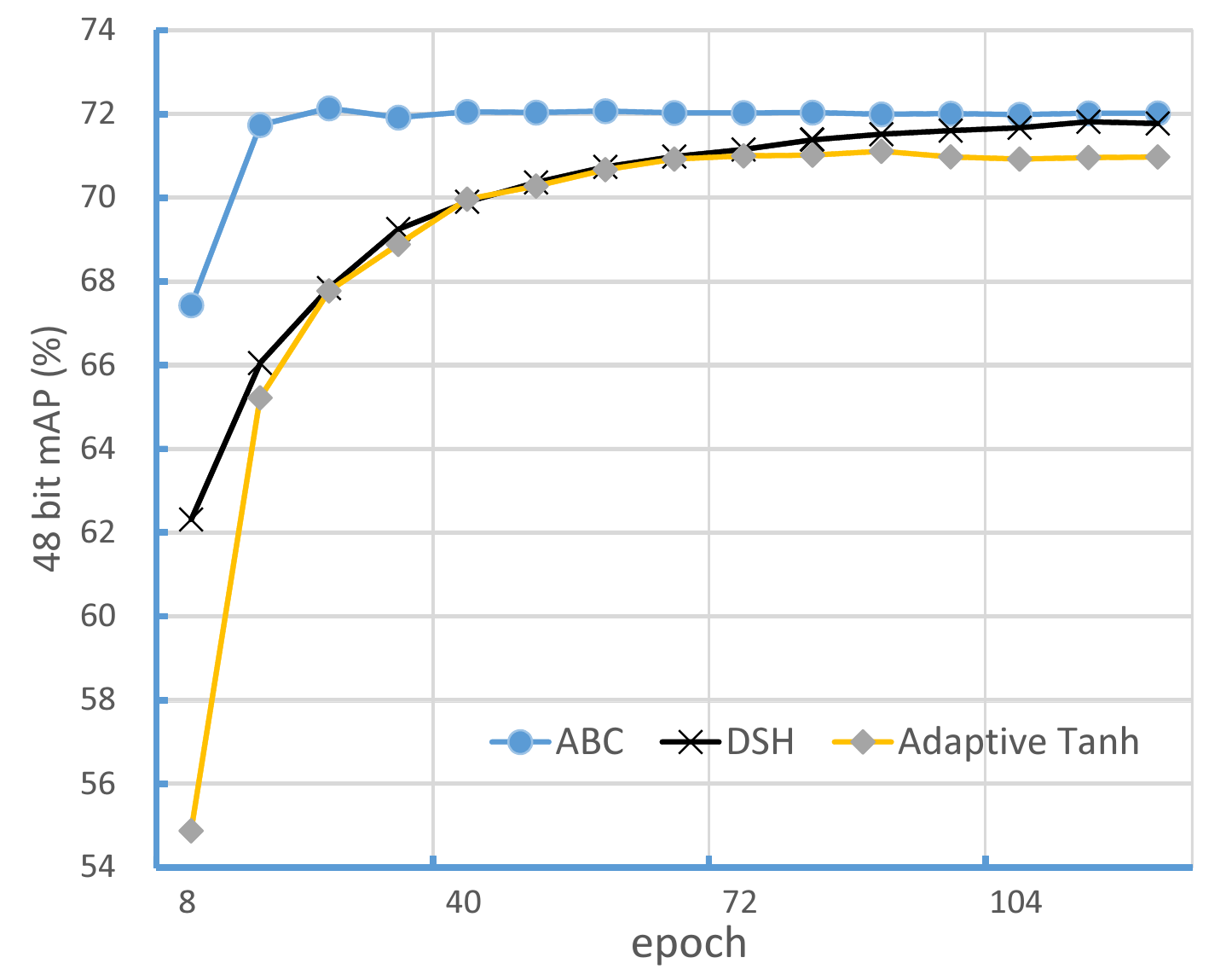} \label{fig:convergence_Cifar}}
	\\
	\subfloat[CIFAR-10 retrieval 12 bits (the higher the better)] { \includegraphics[width=0.34\textwidth]{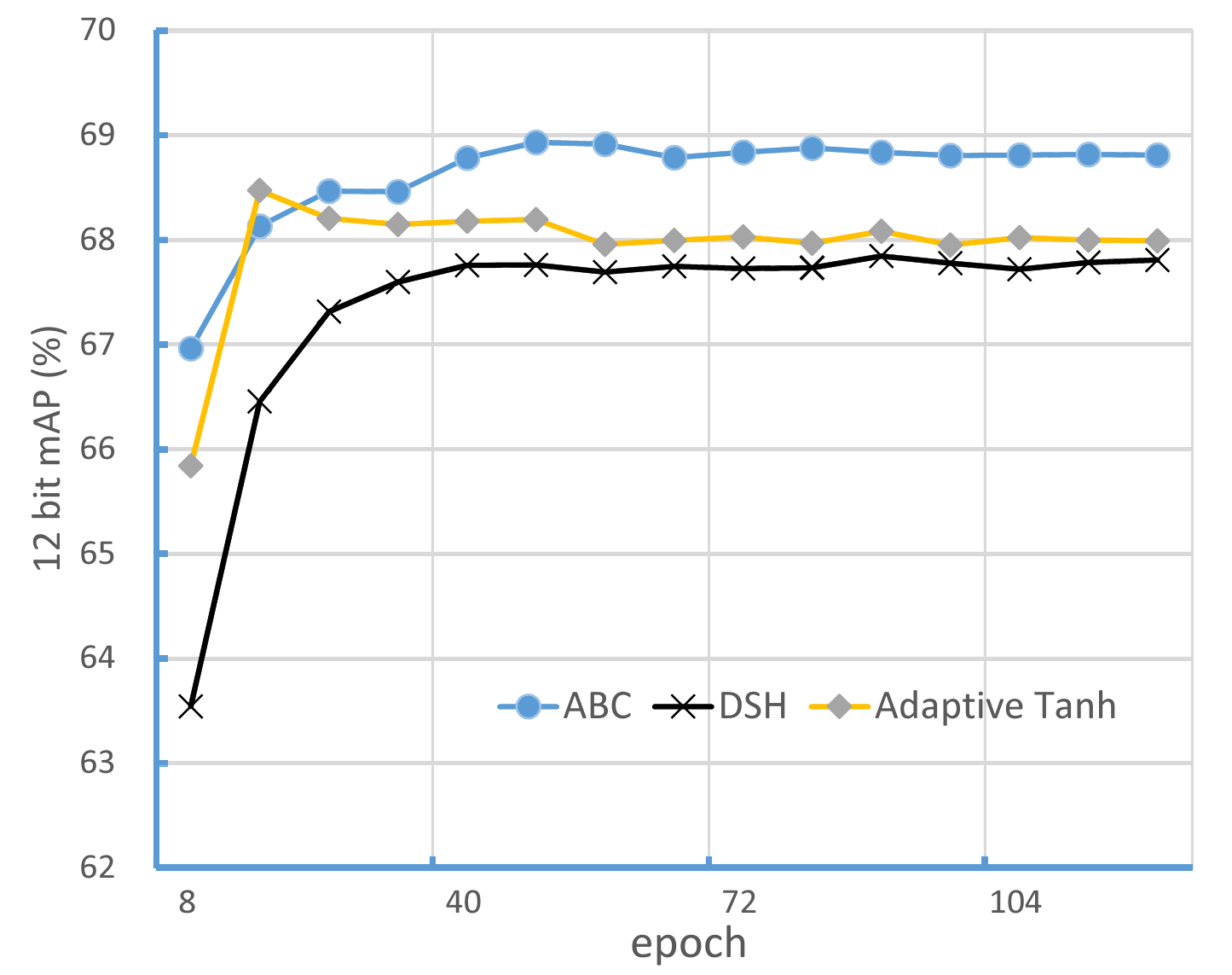} \label{fig:convergence_Cifar_12}}
	\caption{Convergence comparison between the proposed ABC activation function and the adaptive $\tanh$ function.} \label{fig:convergence}
\end{figure*}

Furthermore, we argue that one related benefit of ABC is its fast convergence speed, as shown in Fig.~\ref{fig:convergence}. Let us first observe Fig.~\ref{fig:convergence_ImageNet}, which shows the \emph{validation} set top-1 error for the ImageNet fine-tuning process. After the first epoch finishes, ABC is roughly 4\% better than adaptive $\tanh$. Note that the final difference between these two models is $26.430-24.316=2.114\%$, ABC already converges faster than adaptive $\tanh$ in the first epoch. The difference in convergence speed is clearer when we observe longer. ABC almost converges after 4 epochs, while adaptive $\tanh$ tend to converge after 10 epochs.

Fig.~\ref{fig:convergence_Cifar} shows the convergence comparison for ABC, DSH and adaptive $\tanh$ on the CIFAR-10 image retrieval task, in which 48 bits are used. Note that BN is used in both DSH and adaptive $\tanh$ for these results in Fig.~\ref{fig:convergence_Cifar}, because under this setting all three methods have similar final mAP rates, which facilitates our comparison at earlier epochs. All results are retrieval mAP on the \emph{test} set.

After 8 epochs, ABC's test set retrieval mAP is roughly 5\% higher than that of DSH, but more than 12.5\% higher than that of adaptive $\tanh$. This observation seems to strongly support that saturated functions such as $\tanh$ is not friendly to deep neural network optimization. 

The overall convergence plot again support this claim. ABC almost converges after only 16 epochs, while the convergence epoch index for DSH is 104, and for adaptive $\tanh$ the convergence threshold is 72. The slow convergence speed of DSH might be attributed to how it achieves binary outputs, which are implicitly encouraged by an additional loss term. This implicit form of binarization is inferior to explicit binary constraints in adaptive $\tanh$ (by increasing $\alpha$) and in ABC (by decreasing $r$ till 0). It is not surprising that ABC has the best convergence behavior, because it has explicit binarization forces and is non-saturating.

Another interesting convergence behavior is shown in Fig.~\ref{fig:convergence_Cifar_12}, for 12 bit code length on the CIFAR-10 dataset. In this setup, adaptive $\tanh$ reaches its highest accuracy very fast (after 16 epochs). But, as more epochs are carried out, its accuracy gradually and slightly reduces. Although adaptive $\tanh$ may be unstable for short code length, ABC is quite stable and converges fast.

\subsection{Generalizing to object detection} \label{sec:exp:detection}

In this section, we show that the two ABC models learned by fine-tuning ResNet-50 on ImageNet can generalize well to other vision tasks, in particular object detection, which motivates our design of binary visual representations (cf. Fig.~\ref{fig:motivation}).

We name the two ABC versions as ABC-2048 and ABC-4096, respectively, depending on the length of binary bits. We apply these ABC models in Fast R-CNN~\cite{fastrcnn} to observe how the learned binary representations perform in new tasks. We implement the detection model using Torch. In the training stage, we use selective search's quality mode to extract 2,000 proposals per image (the same in the inference). For ABC-2048, ABC-4096 and ResNet-50, we add a RoI-pooling~\cite{fastrcnn} layer right before {\em res\_{5a}}, on the top of which we also append fully-connected layers from these pretrained models. Weight decay is set to 0.0005 and momentum is 0.9. The initial learning rate is 0.001 and is divided by 10 every 200 epochs.

We compare ABC-2048 and ABC-4096 based detectors with the detector based on the baseline ResNet-50 pretrained model. Because all other factors in the experiments are the same, the difference in detection mAP should reflect different generalization ability of these pretrained models. Note that we set $r=0.1$ for ABC, \emph{i.e.,} not requiring true binary in detection. However, representations are close to binary in both ABC-2048 and ABC-4096.

We train on VOC07 trainval and test on VOC07 test~\cite{voc07}. The Torch pretrained ResNet-50 achieves 68.1\% mAP, while ABC-2048 and ABC-4096 has 68.6\% and 69.1\%, respectively. That is, by replacing real-valued ResNet-50 with approximately binary representations, ABC-2048 and ABC-4096 increased the detection by 0.5\% and 1\%, respectively. This result corroborates our motivation in Fig.~\ref{fig:motivation} that feature magnitudes are not as important as signs (activated or not). In contrast, if we replace the ABC activation function of these two models (ABC-2048 and ABC-4096) with standard ReLU function, the mAP values are only 68.0\% and 68.2\%, lower than our ABC models. Hence, true or approximately binary visual representations such as those produced by ABC could be attractive beyond the retrieval domain.

\section{Conclusions}

We proposed Approximately Binary Clamping (ABC), an activation function and layer for CNN learning, in order to emit binary visual representations. Different from hashing methods that are mainly for reducing computational and storage costs in relatively small scale vision tasks, the motivation and objective for ABC is that  binary visual features may have better generalization in large scale vision tasks.

ABC is a non-saturating, computationally very cheap, easy to implement method. Although traditional hashing methods converge slowly and emit only approximately binary values, ABC converges fast and can output either approximately or true binary values. To the best of our knowledge, ABC is the first binary visual representation that achieves comparable accuracy with real-valued networks. ABC generalized better than regular networks in object detection, and achieved state-of-the-art results in image retrieval.

There are interesting future directions to explore, such as ABC for multi-label learning and to binarize more layers.

\clearpage

\bibliographystyle{splncs}
\bibliography{egbib}
\end{document}